\documentclass[aip, jap, amsmath, amssymb, reprint]{revtex4-1}

\usepackage{natbib}
\usepackage{graphicx}
\usepackage{dcolumn}
\usepackage{bm}

\usepackage[utf8]{inputenc}
\usepackage[T1]{fontenc}
\usepackage{mathptmx}
\usepackage{etoolbox}
 \usepackage{helvet} 
\usepackage{tikz}
\usetikzlibrary{shapes.geometric, arrows, arrows.meta, positioning,fit,backgrounds, shadows}
\makeatletter
\def\@email#1#2{%
 \endgroup
 \patchcmd{\titleblock@produce}
  {\frontmatter@RRAPformat}
  {\frontmatter@RRAPformat{\produce@RRAP{*#1\href{mailto:#2}{#2}}}\frontmatter@RRAPformat}
  {}{}
}%
\makeatother
\begin{document}

\preprint{AIP/123-QED}

\title[Critical Phenomena in Graph Reasoning]{Self-Organizing Graph Reasoning Evolves into a Critical State for Continuous Discovery Through Structural–Semantic Dynamics
}
\author{Markus J. Buehler}
 \email{mbuehler@MIT.EDU.}
\affiliation{Massachusetts Institute of Technology, 77 Mass. Ave, Cambridge, MA 01921, USA. 
}
\date{\today}

\begin{abstract} 
 
We report fundamental insights into how agentic graph reasoning systems spontaneously evolve toward a critical state that sustains continuous semantic discovery. By rigorously analyzing structural (Von Neumann graph entropy) and semantic (embedding) entropy, we identify a subtle yet robust regime in which semantic entropy persistently dominates over structural entropy. This interplay is quantified by a dimensionless Critical Discovery Parameter ($\mathcal{D}$), which stabilizes at a small negative value ($\mathcal{D} \approx -0.03$), indicating a consistent excess of semantic entropy. Empirically, we observe a stable fraction ($\sim 12\%$) of ``surprising'' edges---links between semantically distant concepts---providing evidence of long-range or cross-domain connections that drive continuous innovation. Concomitantly, the system exhibits scale-free and small-world topological features, alongside a negative cross-correlation between structural and semantic measures, reinforcing the analogy to self-organized criticality. These results establish clear parallels with critical phenomena in physical, biological, and cognitive complex systems, revealing an entropy-based principle governing adaptability and continuous innovation. Crucially, semantic richness emerges as the underlying driver of sustained exploration, despite not being explicitly used by the reasoning process. Our findings provide interdisciplinary insights and practical strategies for engineering intelligent systems with intrinsic capacities for long-term discovery and adaptation, and offer insights into how model training strategies can be developed that reinforce critical discovery.

\end{abstract}

\maketitle

\section{\label{sec:level1}Introduction}

Generative modeling for language, vision and other modalities has received significant interest in the past years~\cite{Vaswani2017AttentionNeed_fixed,AlecRadfordImprovingPre-Training,li2023textbooksneediiphi15,dubey2024llama3herdmodels,abdin2024phi3technicalreporthighly,Brown2020LanguageLearners,salinas2025exoplanettransitcandidateidentification,abdin2024phi4technicalreport}, including the development of reasoning models that exhaust thinking and reflection strategies before responding to tasks~\cite{AlQuraishi2019End-to-EndStructure,zelikman2022starbootstrappingreasoningreasoning,buehler2024preflexorpreferencebasedrecursivelanguage}. However, little is known about the mechanisms by which models, especially reasoning models, develop answers and whether general principles can be extracted.


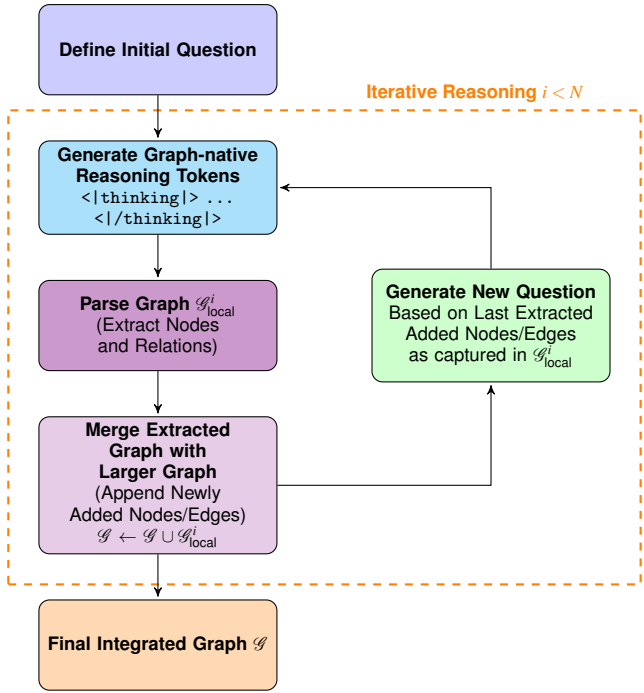
\begin{figure}
\centering

\sffamily
\scriptsize


\begin{tikzpicture}[
    node distance=0.6cm and 2cm, auto,
    task/.style={rectangle, draw, fill=blue!20, text width=3cm, text centered, rounded corners, minimum height=1.2cm},
    generation/.style={rectangle, draw, fill=cyan!30, text width=3cm, text centered, rounded corners, minimum height=1.2cm},
    extraction/.style={rectangle, draw, fill=violet!40, text width=3cm, text centered, rounded corners, minimum height=1.2cm},
    merge_graph/.style={rectangle, draw, fill=violet!20, text width=3cm, text centered, rounded corners, minimum height=1.5cm},
    question/.style={rectangle, draw, fill=green!20, text width=3cm, text centered, rounded corners, minimum height=1.5cm},
    visualization/.style={rectangle, draw, fill=yellow!30, text width=3cm, text centered, rounded corners, minimum height=1.2cm},
    final_output/.style={rectangle, draw, fill=orange!30, text width=3cm, text centered, rounded corners, minimum height=1.2cm},
    line/.style={draw, -stealth', shorten >=1pt},
    dashed line/.style={draw, dashed, shorten >=1pt}]

    \node [task] (start) {\textbf{Define Initial Question} };
    \node [generation, below=of start] (generate) {\textbf{Generate Graph-native \\ Reasoning Tokens} \\ \texttt{<|thinking|> ... <|/thinking|>}};
    \node [extraction, below=of generate] (extract_graph) {\textbf{Parse Graph} \( \mathcal{G}_{\text{local}}^i \) \\ (Extract Nodes and Relations)};
    \node [merge_graph, below=of extract_graph] (merge_graph) {\textbf{Merge Extracted Graph with \\ Larger Graph} \\ (Append Newly Added Nodes/Edges) \\  $\mathcal{G} \leftarrow \mathcal{G} \cup \mathcal{G}_{\text{local}}^i$};
    \node [final_output, below=of merge_graph] (final_result) {\textbf{Final Integrated Graph} $\mathcal{G}$} ;

    \path [line] (start) -- (generate);
    \path [line] (generate) -- (extract_graph);
    \path [line] (extract_graph) -- (merge_graph);
    \path [line] (merge_graph) --  (final_result);

    \node [question, right=of extract_graph, xshift=-.75cm] (new_question) {\textbf{Generate New Question} \\ 
    Based on Last Extracted Added Nodes/Edges as captured in $\mathcal{G}_{\text{local}}^i$ };
    \path [line] (merge_graph.east) -| (new_question.south);
    \path [line] (new_question.north) |- (generate.east);

    \begin{pgfonlayer}{background}
        \node [draw=orange, thick, dashed, fit=(generate) (extract_graph) (merge_graph) (merge_graph) (new_question), 
               inner sep=0.4cm, label={[text=orange]above, xshift=2cm: \textbf{Iterative Reasoning} $i<N$}] (iteration_box) {};
    \end{pgfonlayer}

\end{tikzpicture}

\rmfamily
\caption{Algorithm used for iterative knowledge extraction and graph refinement as reported in~\cite{buehler2025agenticdeepgraphreasoning}. 
At each iteration \( i \), the model generates reasoning tokens that include a graph representation of the thinking process (blue). 
A local graph \( \mathcal{G}_{\text{local}}^i \) is then extracted (violet) and merged with the global graph \( \mathcal{G} \) (light violet). A follow-up task is then generated based on the latest extracted nodes and edges in \( \mathcal{G}_{\text{local}}^i \) (green), leading to iterative reasoning (orange), so that the model expands the graph with increasing number of nodes and edges. }
\label{fig:fig_01_earlier}
\end{figure}

One particular class of reasoning models, agentic deep graph reasoning models such as Graph-PRefLexOR, iteratively construct knowledge graphs by recursively applying neural reasoning over extended test-time compute~\cite{brown2020language,yao2023tree,buehler2024preflexorpreferencebasedrecursivelanguage,buehler2025insitugraphreasoningknowledge}. While previous work has established the overall capability of such graph-native reasoning models, the fundamental physical principles governing their structural and semantic evolution remain largely unexplored, albeit earlier work has proposed graph-focused strategies that also incorporate category theory~\cite{buehler2024preflexorpreferencebasedrecursivelanguage,Buehler2025GraphAwareGPT,buehler2025agenticdeepgraphreasoning,Spivak2011CategoryNetworks,Giesa2011ReoccurringAnalogies,Giesa2012CategoryDesign}. The structured generation of thinking mechanisms offers the potential to conduct more rigorous analyses of the resulting graph structures. Relatedly, research indicates that standard Transformer architectures can be interpreted as a variant of the Graph Isomorphism Network (GIN), where attention mechanisms function over relational structures rather than purely sequential token representations~\cite{Buehler2025GraphAwareGPT}. These relations of AI models with established concepts from mathematics and physics provide grounds for a wider-ranging analysis of their behavior through a lens of dynamical systems.

\begin{figure*}
\includegraphics[width=1.\linewidth]{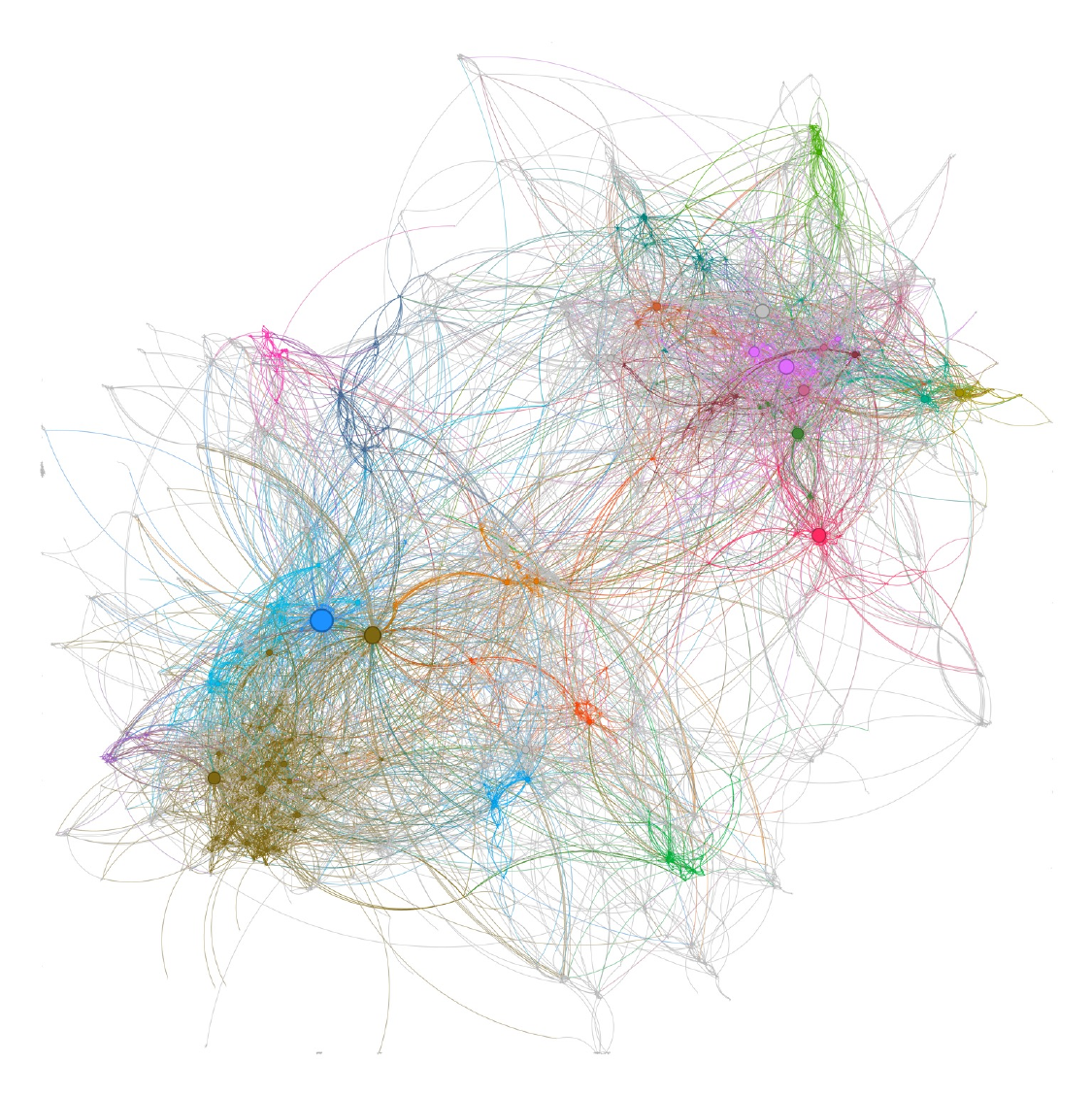}
\caption{\label{fig:fig_0} The complete knowledge graph generated by the agentic deep graph reasoning model (Graph-PRefLexOR~\cite{Buehler2025GraphAwareGPT}) after iterative evolution. Nodes and edges emerge iteratively through recursive reasoning, forming complex structural communities. Colors reflect different communities (up to 20 unique communities shown).}
\end{figure*}

The work reported here uses the series of graph networks obtained in earlier work, specifically~\cite{buehler2025agenticdeepgraphreasoning} (see, Materials and Methods for details, and particularly the original paper, where the algorithm summarized in Figure~\ref{fig:fig_01_earlier} is explained in further detail). As shown in the figure, the graphs were obtained through an iterative, agentic process where a reasoning-native large language model autonomously expanded and refined a knowledge graph. At each step, the system generated new concepts and relationships, integrated them into the graph, and formulated subsequent prompts based on the evolving structure. This resulted in hundreds of graphs that allow us to study their detailed evolution as the reasoning process expands. The earlier work~\cite{buehler2025agenticdeepgraphreasoning} has shown notable properties of these graphs, such as that it resulted in a scale-free network with emergent hubs and bridges linking disparate knowledge clusters~\cite{stanley1987introduction,barabasi1999emergence}. Over hundreds of iterations, new nodes and edges continuously appeared, centrality measures evolved, and shortest path distributions adapted, leading to increasingly distributed connectivity. 

\begin{figure}
\includegraphics[width=.75\linewidth]{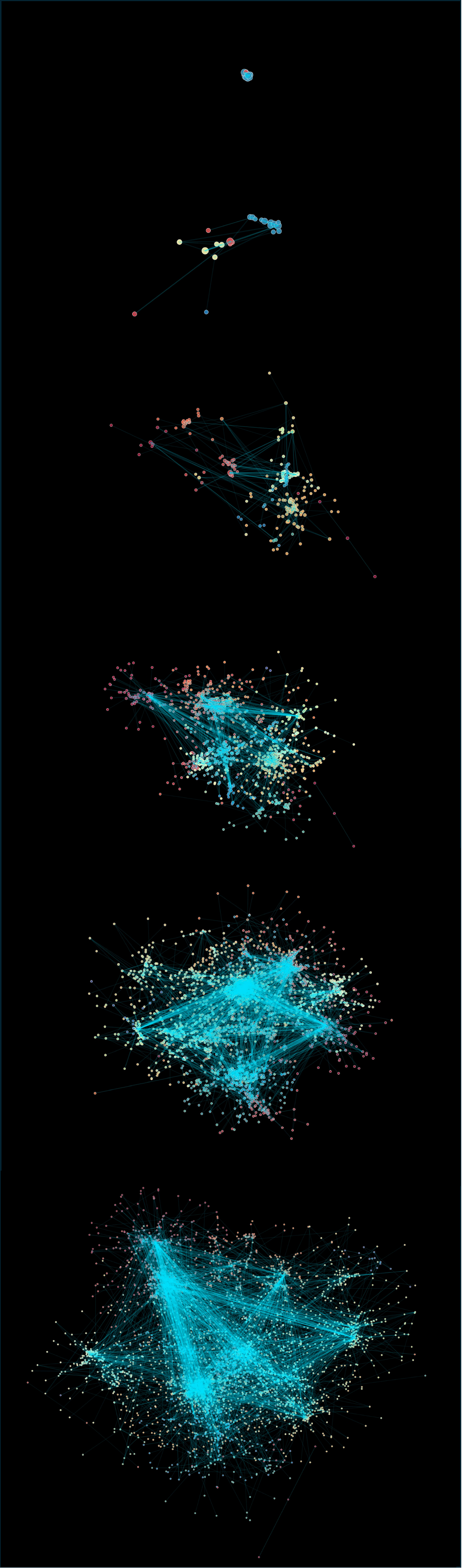}
\caption{\label{fig:fig_10} Growth of the network over reasoning iterations, three-dimensional view. The evolution of the network over reasoning iterations is clearly visible. }
\end{figure}

Fig.~\ref{fig:fig_0} shows a depiction of the network generated at the end of the reasoning iterations~\cite{buehler2024preflexorpreferencebasedrecursivelanguage,Buehler2025GraphAwareGPT,buehler2025agenticdeepgraphreasoning}. 
Complementing the two-dimensional projection in Fig.~\ref{fig:fig_0}, Fig.~\ref{fig:fig_10} shows the growth of the network over reasoning iterations in a three-dimensional view. We note that a wealth of analysis is included in the earlier work~\cite{Buehler2025GraphAwareGPT,buehler2025agenticdeepgraphreasoning}, on top of which this study is built. We refer readers to the original papers for further details, albeit key results will be reviewed here for clarity.

In this paper, we present insights revealing how agentic graph reasoning systems spontaneously evolve behaviors analogous to critical phenomena observed in physical, biological, and cognitive complex systems~\cite{stanley1987introduction,kauffman1993origins}. Through rigorous quantitative analysis of structural entropy (Von Neumann graph entropy) and semantic embedding entropy we posit that a graph representing an evolving knowledge system has two parallel dimensions: network topology and conceptual diversity. Hence, complementing Von Neumann graph entropy, we define semantic entropy as a measure quantifying how conceptually diverse or spread out the node representations are within a learned embedding space, computed via the spectral properties of a similarity (cosine-based) adjacency matrix derived from pretrained language model embeddings. These measures of entropy serve as analogues to physical entropy measures. This concept builds on earlier work~\cite{CarnapBarHillel1953} that introduced an early, formal approach to semantic information rooted in logical probability and content, establishing a foundation for later entropy-based interpretations of meaning. Other research~\cite{Mikolov2013} introduced word embedding techniques that underpin contemporary distributional semantics, providing the vector-space basis for the concept of semantic adjacency construction. Related, information-theoretic tools~\cite{SoleValverde2004} were applied to complex networks, illustrating how entropy concepts elucidate architectural constraints and evolutionary dynamics, a perspective we extend by incorporating semantic embeddings.

\begin{figure}
\includegraphics[width=.9\linewidth]{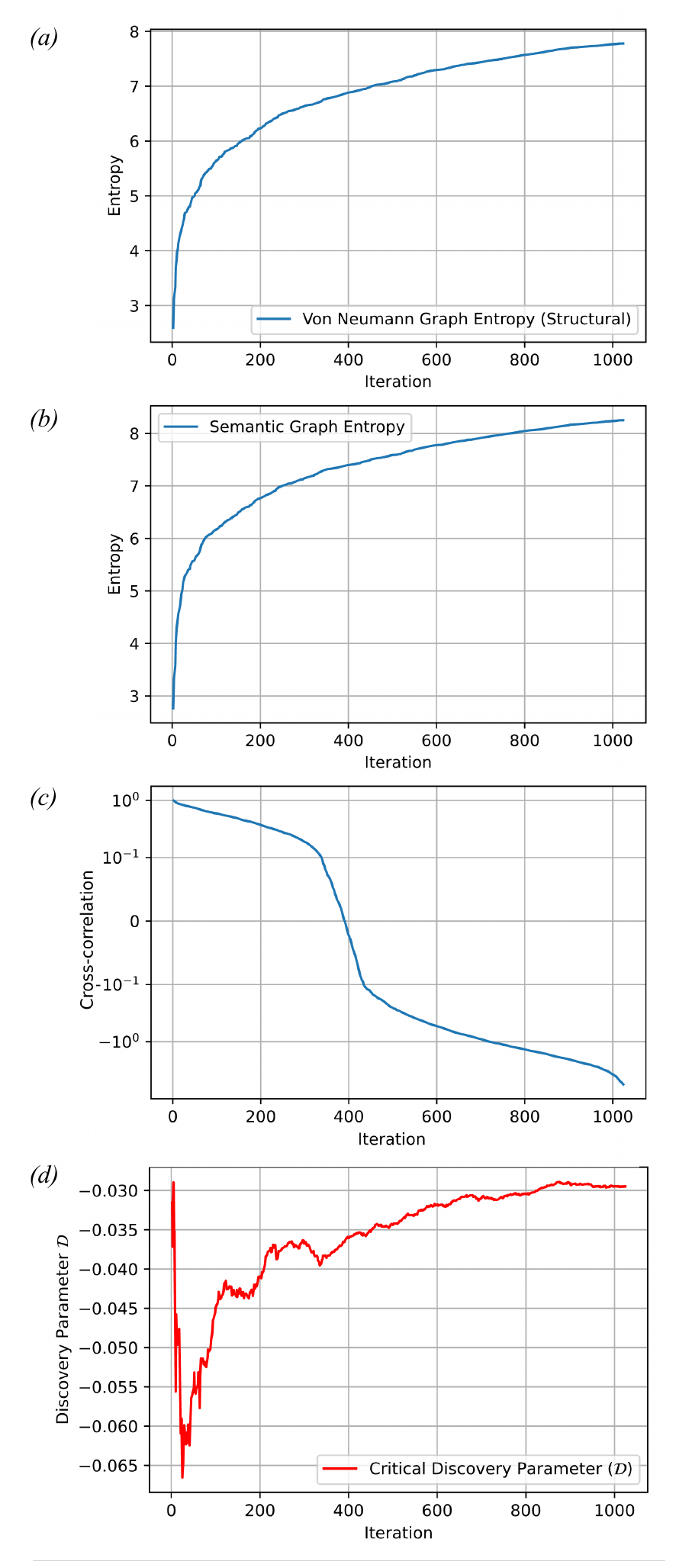}
\caption{\label{fig:fig_1} Evolution and comparative analysis of structural and semantic entropy during graph reasoning. (a) Structural entropy, quantified by Von Neumann Graph Entropy, increases rapidly initially and stabilizes gradually, indicating consistent structural complexity growth. (b) Semantic entropy evolves similarly but remains consistently higher than structural entropy, indicating sustained semantic complexity dominance.
(c) Cross-correlation between structural and semantic entropy reveals a critical transition near iteration 400, shifting from positively correlated (co-evolution) to negatively correlated dynamics (semantic-structural divergence), reminiscent of a phase transition.(d) The Critical Discovery Parameter ($\mathcal{D}$) stabilizes at a slightly negative value ($\mathcal{D}\approx -0.03$), explicitly confirming persistent semantic entropy dominance and guiding structural evolution towards sustained exploratory innovation. Together, these results explicitly demonstrate that semantic dynamics consistently lead and shape structural evolution, underpinning continuous semantic exploration and innovation.}
\end{figure}

\section{Results and Discussion}

Our analysis is conducted over the growing graphs that emerge during a lengthy reasoning process. 
Fig.~\ref{fig:fig_1}(a) shows that both structural (Von Neumann graph entropy) and semantic entropy rapidly increase and stabilize over the course of iterative reasoning cycles, clearly indicating continuous growth in both structural and semantic complexity. However, Fig.~\ref{fig:fig_1}(b) reveals that semantic entropy consistently remains higher than structural entropy throughout all iterations, explicitly signifying sustained semantic dominance in the network's evolution. Taken together, the data shows that the system self-organizes to maximize informational entropy in its knowledge network.

\begin{figure}
\includegraphics[width=.9\linewidth]{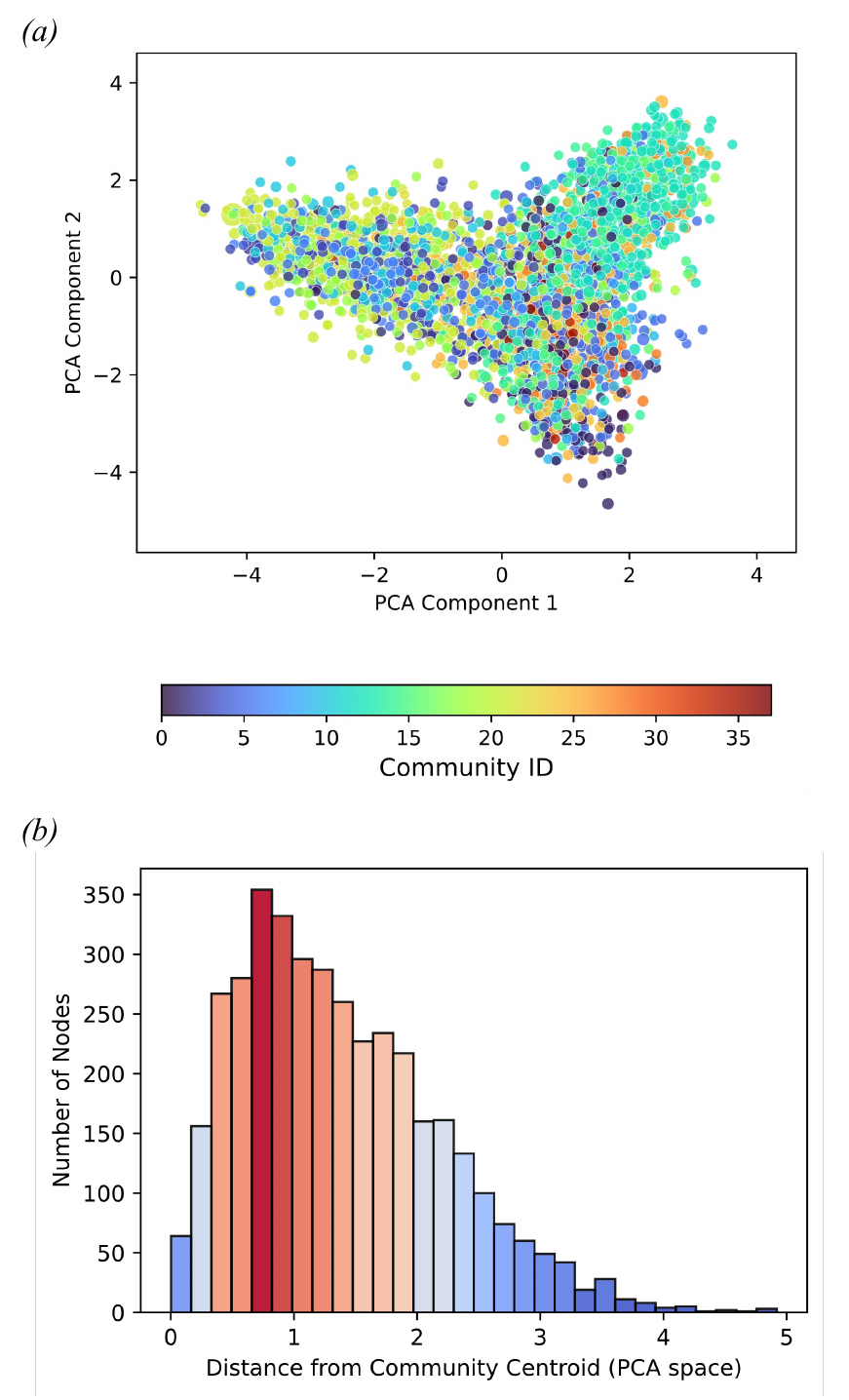}
\caption{\label{fig:fig_5} Semantic Embedding Space and Structural Community Decoupling. (a) Two-dimensional PCA projection of semantic node embeddings, colored according to structural communities identified via the Louvain method. The observed partial mixing of community colors in semantic space demonstrates a critical decoupling between semantic embeddings and structural clusters, highlighting complementary yet distinct forms of knowledge encoded within the evolving graph. It confirms that the system’s structural connections are not dictated simply by semantic proximity. (b) Histogram of node distances from their respective community centroids in PCA space. The $x$-axis represents the distance from the centroid, while the $y$-axis denotes the number of nodes at each distance.  The color gradient reflects node density, with red indicating higher counts and blue representing lower values. The distribution is right-skewed and long-tailed, with most nodes clustered around a distance of $1$, while a few nodes exhibit significantly larger distances. }
\end{figure}

To more deeply understand this interplay, we computed the cross-correlation between these entropies (Fig.~\ref{fig:fig_1}(c)), explicitly demonstrating a clear critical transition near iteration 400, where structural-semantic correlation shifts from positive (initially co-evolving structure and semantics) to strongly negative values. This negative correlation explicitly indicates that structural decisions increasingly diverge from underlying semantic relationships, systematically exploring structurally justified yet semantically novel connections.

Finally, Fig.~\ref{fig:fig_5}(a) shows a two-dimensional PCA projection of semantic embeddings, with nodes colored by their Louvain structural community memberships. Notably, structural communities are not distinctly separated in semantic embedding space; instead, they appear intermingled and partially overlapping. This partial decoupling between structural clusters and semantic similarity demonstrates that the knowledge graphs produced by the reasoning model encode structural and semantic information through fundamentally distinct but complementary dimensions. Such a result reveals how the co-existence and partial independence of semantic and structural information allow the system to remain simultaneously robust (via structural organization) and flexible (via semantic novelty). This is because the Louvain algorithm discovers communities based purely on the graph’s adjacency. Meanwhile, semantic embeddings represent conceptual similarity. Because we see multiple structural communities overlapping or interspersed in embedding space, the plot directly visualizes the partial semantic–structural divergence that our earlier metrics indicated.  Fig.~\ref{fig:fig_5}(b) shows a distance histogram, whose skewed distribution confirms that while structural communities have some semantic coherence, they also contain outlier or bridging nodes. 

Taken together, these results reveal important connections between physical concepts of entropy, critical phase transitions, and continuous discovery processes in artificial intelligence reasoning processes. They establish clear parallels between emergent graph reasoning and physical systems, suggesting novel interdisciplinary approaches to engineer intelligent, adaptive reasoning systems informed by principles from statistical physics and complex system theory, whereby the graph reasoning process spontaneously evolves toward a ``critical state''  that sustains ongoing discovery.

\subsection{Unified Concepts of Critical Discovery}

Based on our observations, we propose a unified theory, the critical discovery principle, which succinctly captures the inherent dynamics of agentic reasoning systems by quantifying the relative dominance of structural entropy compared to semantic entropy. The central premise is that autonomous reasoning processes spontaneously evolve toward and sustain a critical balance between structural complexity and semantic novelty. This balance, where the system maintains itself near criticality without external tuning, is quantified by a dimensionless discovery parameter defined as:
\[
\mathcal{D} = \frac{S_{\text{struct}} - S_{\text{sem}}}{S_{\text{struct}} + S_{\text{sem}}},
\]
where \( S_{\text{struct}} \) and \( S_{\text{sem}} \) denote structural and semantic entropies, respectively.

In our numerical analysis of the results, the Critical Discovery Parameter ($\mathcal{D}$) explicitly quantifies this balance between structural and semantic entropy (Fig.~\ref{fig:fig_1}(d)). The parameter stabilizes near a small negative value ($\mathcal{D}\approx -0.03$), explicitly confirming that semantic entropy subtly dominates structural entropy, mirroring the competition between energy and entropy in phase transitions. This is reminiscent of a second-order phase transition: initially, structure and content co-evolve (like an order parameter following the external field), but beyond the critical point, their relationship inverts – similar to how beyond the Curie temperature, magnetization collapses even as thermal fluctuations (entropy) continue to increase.

\begin{figure}
\includegraphics[width=0.9\linewidth]{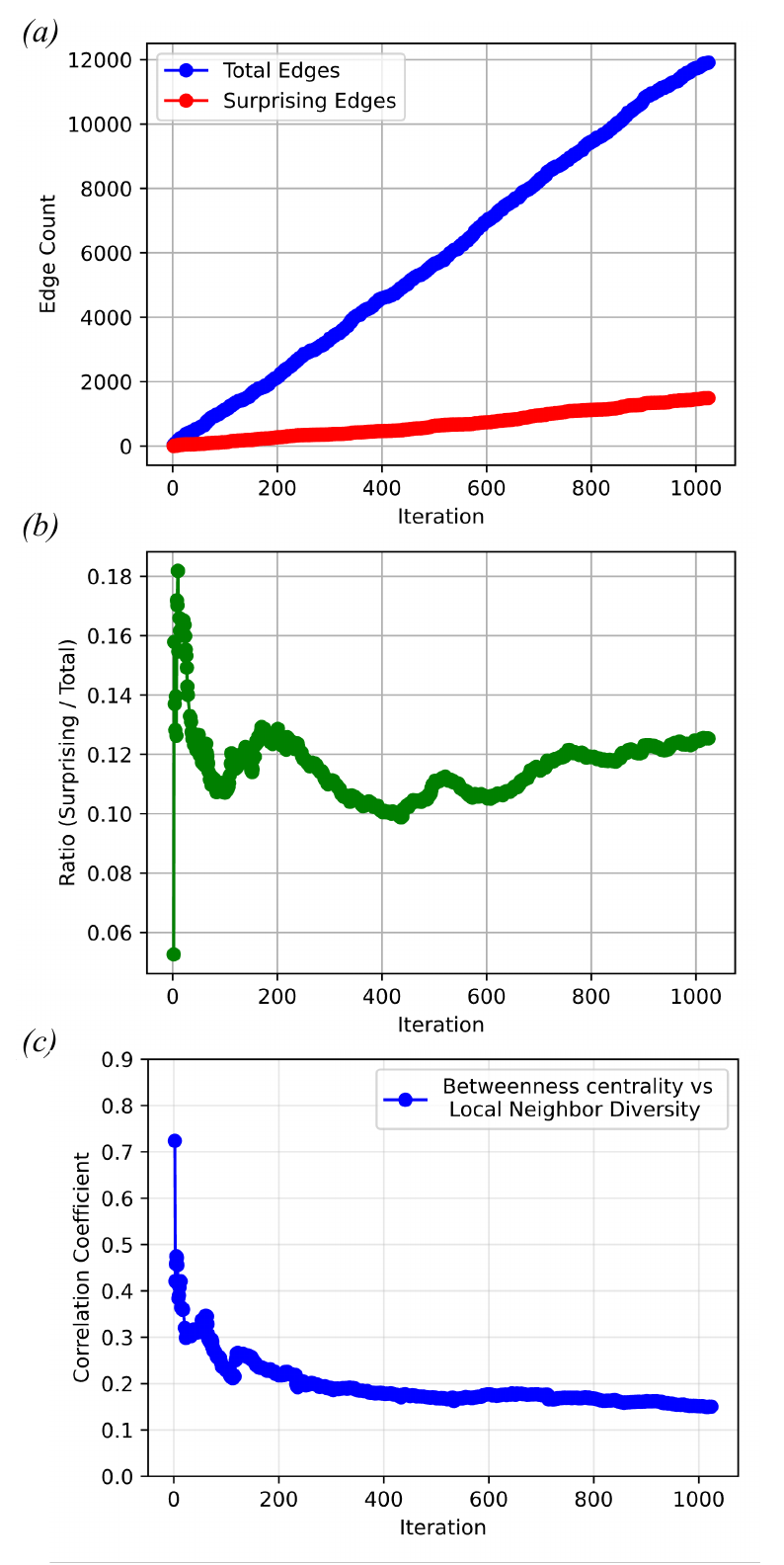}
\caption{\label{fig:fig_3}Analysis of surprising edges as a measure of continuous discovery. (a) Evolution of total and surprising edge counts; (b) The stable fraction ($\sim12\%$) of surprising edges indicates sustained semantic exploration and discovery capacity. (c) Correlation between betweenness centrality and local semantic neighbor diversity. After large values in small and early-stage graphs (aggressive initial phase of semantic bridging), the correlation stabilizes at a positive value, indicating that nodes on many shortest paths connect neighbors with more diverse semantic embeddings.}

\end{figure}

To empirically validate this theory, we explicitly analyze the evolution of surprising edges—edges that are structurally connected but semantically distant—in Fig.~\ref{fig:fig_3}(a). Although the total edge count continuously increases, the ratio of surprising edges
\[
\frac{N_s}{N} \rightarrow \alpha,
\]
stabilizes around $\alpha \approx 12\%$ (Fig.~\ref{fig:fig_3}(b)). This explicitly indicates a sustained intrinsic mechanism for semantic exploration and innovation. This stable fraction explicitly characterizes a subtle balance: semantic entropy consistently guides structural evolution toward novel relationships without allowing the system to descend into disorder or rigidity. Such behavior closely mirrors critical phenomena observed in physical and biological systems, which similarly exhibit a delicate equilibrium between predictable structural order and adaptive semantic novelty. The persistent fraction $\alpha$ thus explicitly emerges as an empirical hallmark of criticality, analogous to residual disorder found at critical points in physical systems. Interestingly, the fraction of cross-domain connections aligns with magnitudes seen in small-world networks needed to drastically reduce path lengths~\cite{watts1998collective,PhysRevLett.87.198701}. This hints that the reasoning process naturally populates the graph with just enough long-range links to hit a connectivity threshold, ensuring information can propagate across the entire network efficiently. This nuanced yet robust semantic dominance explicitly validates the Critical Discovery Principle, highlighting semantic complexity as the primary driver behind continuous exploratory innovation.

To further elucidate the structural-semantic dynamics, we computed the correlation between betweenness centrality (BC) and  local semantic neighbor diversity  (Fig.~\ref{fig:fig_3}(c)). In the early iterations, the correlation fluctuates significantly as the network is small or changing quickly, with early high positive correlation indicating an aggressive initial phase of semantic bridging. As the graph grows the correlation settles at a positive level, saturating at around 400 iterations. This suggests that high-BC nodes, which act as bridges, tend to have neighbors that are more spread out in embedding space. The network consistently maintains moderate semantic diversity around central nodes, providing structural stability and continuous innovation through balanced semantic exploration.

In summary, our theory captures the subtle yet persistent semantic dominance and sustained exploratory capability observed experimentally. The stable fraction of structurally coherent yet semantically distant (``surprising'') edges explicitly serves as empirical validation of the Critical Discovery Principle, highlighting both its predictive power and potentially a broad conceptual relevance across artificial intelligence, physics, and complexity science.

\subsection{Interdisciplinary Parallels and Universality}

The Critical Discovery Principle identified in our agentic graph reasoning model closely mirrors phenomena observed across diverse natural and artificial complex systems. In biological networks, gene regulatory and protein interaction networks maintain stability through structurally coherent interactions explicitly guided by underlying functional (semantic) requirements, continuously introducing exploratory variations essential for evolutionary adaptation. Similarly, neural and cognitive systems exhibit structural-functional coupling near critical states (e.g., neuronal avalanches), promoting cognitive flexibility and innovation through persistent semantic enrichment. Ecological systems analogously balance stable structural interactions among species with exploratory semantic dynamics driven by migration, mutation, and invasion, ensuring resilience and adaptability.

In social and economic contexts, innovation networks naturally balance structural coherence—stable technological or economic sectors—with continuous semantic-driven exploration of novel, disruptive ideas essential for sustained growth. Physical and material systems undergoing second-order phase transitions (e.g., magnetic, superconducting, or critical fluid systems) also display analogous behaviors, characterized by subtle structural-semantic interplay at critical boundaries. Similarly, glassy and amorphous materials maintain residual structural defects at criticality, explicitly providing mechanical adaptability reminiscent of our model's structurally coherent yet semantically novel (``surprising'') edges. Finally, artificial intelligence systems, particularly reinforcement learning frameworks, inherently balance exploitation of structurally known knowledge with exploration driven by semantically novel opportunities, embodying the universal criticality we explicitly describe.

These interdisciplinary parallels explicitly suggest that the subtle yet persistent semantic dominance and critical discovery dynamics observed in agentic reasoning systems are not isolated phenomena, but rather manifestations of a deep, universal organizing principle governing adaptability, innovation, and discovery across complex adaptive systems. Recognizing such universality explicitly provides foundational insights for interdisciplinary theory and practical strategies to engineer robust, semantically-driven, innovative, and adaptive intelligent systems.

\subsection{Prediction: Emergent Behavior in Novel Reasoning Scenarios}

Drawing inspiration from hierarchical organization observed in biological and physical complex systems operating near criticality, we predict that when the reasoning model encounters significantly larger-scale or multi-layered conceptual problems—tasks fundamentally differing in complexity or dimensionality from previously encountered scenarios—it will spontaneously generate hierarchical structures and multi-scale conceptual organization.

Specifically, analogous to biological networks (e.g., neural circuits, ecosystems) and physical systems exhibiting scale-invariant phenomena near critical points, we anticipate that the reasoning graph will dynamically reorganize into nested communities characterized by clear hierarchies and scale-free connectivity distributions. Structural entropy would initially increase, reflecting new complexity from encountering novel conceptual scales, followed by the spontaneous formation of multi-scale modular structures driven purely by algorithmic structural evolution. Semantic entropy—measured independently—will likely exhibit multiple plateaus, corresponding to semantic stabilization at distinct hierarchical scales of abstraction. We further anticipate that the stable fraction of structurally justified yet semantically distant ("surprising") edges will progressively decrease at finer, local scales, while remaining elevated at global scales, thus ensuring continuous discovery and innovation at higher levels of abstraction.

This prediction reflects known universal properties of critical systems, where introducing entirely new spatial, temporal, or organizational scales spontaneously triggers hierarchical and scale-free pattern formation. Empirically confirming this emergent behavior in future Graph-PRefLexOR experiments would represent compelling evidence of a deeper universal principle underlying complex reasoning and intelligence across diverse scales. As shown in earlier work~\cite{buehler2025agenticdeepgraphreasoning} our system exhibits scale‐free degree distributions, small‐world connectivity, and a stable fraction of semantically distant edges—consistent with classic self‐organized criticality signatures. Together with the near‐balanced but persistent dominance of semantic entropy, these findings align with core self-organized criticality principles: the network’s structure and semantic content spontaneously organize toward a critical point, sustaining continuous novelty while avoiding trivial ordering or random disintegration. 
In fact, it aligns with the notion of self‐organized criticality, where we see a sustained capacity for novelty (semantic) that does not collapse into either random chaos or purely local uniformity. A deeper demonstration of this concept would require further analysis to dig deeper into scale invariance and power‐law behaviors.

\subsection{Future work: Maximizing Discovery via Reinforcement Learning}

Inspired by the critical discovery dynamics identified in this study, we propose a practical reinforcement learning (RL) framework to explicitly maximize the capacity for continuous semantic discovery in agentic graph reasoning systems. Our empirical results suggest that the graph spontaneously stabilizes near a slightly negative critical discovery parameter (\(D \approx -0.03\)), indicative of a subtle but persistent semantic dominance. However, this empirically observed state might not necessarily represent the global optimum. Thus, we introduce a flexible reinforcement learning approach that can systematically guide the system toward optimal conditions for semantic exploration, explicitly encouraging the model to explore richer semantic spaces.
To formalize this as an RL problem, we propose the following reward function:

\begin{equation}
R_t = -\lambda_\mathcal{D} (\mathcal{D}_t - \mathcal{D} _{\text{target}})^2 + \lambda_{SE} S_{\text{sem}}(t) + \lambda_{\alpha}(1 - |\alpha_t - \alpha_{\text{target}}|),
\end{equation}
where:
\begin{itemize}
    \item \(\mathcal{D}  \) quantifies the critical discovery balance between structural and semantic entropy.
    \item \(S_{\text{sem}}(t)\) explicitly measures the semantic entropy, encouraging semantic exploration.
    \item \(\alpha_t\) is the fraction of surprising edges (semantically distant but structurally connected edges).
    \item Hyperparameters \(\lambda_D\), \(\lambda_{SE}\), and \(\lambda_\alpha\) balance the relative importance of each term.
\end{itemize}

Unlike traditional supervised learning, reinforcement learning does not rely on explicit labels. Instead, the model generates actions (such as graph expansions) guided by a learned probability distribution (policy). The gradients flow through this policy, adjusting the parameters to increase the likelihood of actions that yield higher rewards. Formally, the gradient objective (analogous to the gradient of the loss with respect to all model parameters) is:
\begin{equation}
\nabla_\theta J(\theta) = \mathbb{E}_{a \sim \pi_{\theta}(a|G)}\left[R_t \cdot \nabla_{\theta}\log \pi_{\theta}(a|G)\right],
\end{equation}
where:
\begin{itemize}
    \item \(\nabla_\theta J(\theta)\) is the gradient of the expected cumulative reward \(J(\theta)\) with respect to the model parameters \(\theta\).
    \item \(\mathbb{E}\) denotes the expectation, indicating that we average over multiple actions \(a\), which are sampled according to the current policy distribution.
    \item \(a\) represents an action taken by the agent, such as adding a new node or edge in the knowledge graph.
    \item \(R_t\) is the scalar reward associated with action \(a\), explicitly computed from the critical discovery metrics defined earlier. Higher \(R_t\) indicates actions leading to desirable semantic exploration and novelty.
    \item \(\pi_{\theta}(a|G)\) is the policy function parameterized by \(\theta\), representing the probability of taking action \(a\) given the current graph state \(G\). In practice, we use the logarithm of this probability (\(\log \pi_{\theta}(a|G)\)) because it simplifies the numerical computation of gradients and ensures better stability during training.
\end{itemize}
Note that, unlike traditional supervised learning where optimization typically involves minimizing an explicit loss function (such as cross-entropy), reinforcement learning explicitly maximizes an expected reward. Therefore, the gradient defined above naturally appears without a minus sign. However, since standard optimization libraries (e.g., PyTorch) conventionally perform gradient descent (minimizing objectives), practitioners commonly define the reinforcement learning objective with an explicit negative sign as follows:
\[
\mathcal{L}_{\text{RL}}(\theta) = -R_t \cdot \log \pi_{\theta}(a|G).
\]
This negative sign is introduced purely for computational convenience, converting the reward-maximization problem into an equivalent loss-minimization form, aligning with conventional gradient descent routines. Moreover, the theoretical gradient formulation includes an expectation operator \(\mathbb{E}_{a \sim \pi_{\theta}(a|G)}[\cdot]\) because the gradient represents an average over all possible actions according to the policy distribution. Practically, this expectation is approximated by sampling actions from the policy and averaging their resulting rewards.

We note that in reinforcement learning, the policy does not produce explicit answers like in supervised learning. Instead, at every step, it generates a probability distribution over possible actions (such as choosing to add a particular node or edge to the graph). The term \(\log \pi_{\theta}(a|G)\), known as the \textit{log probability}, represents how confident the model was about choosing a specific action \(a\). A high log probability indicates high confidence, while a low log probability indicates uncertainty. During training, actions leading to higher rewards are encouraged by increasing their log probabilities, making these beneficial actions more likely in future predictions. Conversely, actions resulting in low or negative rewards have their log probabilities decreased, reducing their likelihood in future decisions. Thus, gradients flow through these log probabilities, guiding the model towards actions that consistently yield higher rewards and better discovery outcomes.

Intuitively, this equation shows how the gradients ``flow'' during reinforcement learning: the reward \(R_t\) serves as a weighting factor, determining how strongly each action influences parameter updates. Actions yielding higher rewards receive greater weight, making similar actions more probable in future iterations. Conversely, actions resulting in lower or negative rewards reduce their likelihood, guiding the model towards continuous semantic discovery and optimal critical dynamics.

Actions receiving higher rewards become more probable in future iterations, thus shaping the model's semantic discovery behavior.

\section{Conclusions}

This research identifies entropy-based principles governing structural-semantic relationships in artificial reasoning systems. By analyzing entropy dynamics within agentic graph reasoning systems, we uncover insights into the intrinsic nature of continuous discovery and critical phenomena that characterize evolving complex systems. In our experiment, the persistent presence (~12\%) of structurally connected yet semantically distant (``surprising'') edges reveals continuous discovery and adaptive flexibility as emergent properties intrinsic to agentic reasoning models, bridging artificial intelligence, statistical physics, and complex adaptive systems theory. This result confirms the system’s ongoing ability to form structurally significant but conceptually ``far'' connections, thereby operationalizing the idea of semantic ``dominance'' in a measurable way. 

The agentic graph reasoning behaves as a self-organizing critical system, with a critical point as an attractor of its dynamics. The agentic graph reasoning model spontaneously evolves into a critical state, analogous to a high-temperature thermodynamic phase where semantic entropy (favoring disorder) persistently dominates structural organization (favoring order), resulting in a stable, mildly negative critical discovery parameter $\mathcal{D}$ reminiscent of a free-energy minimum shifted toward disorder~\cite{bak1987self}, providing evidence that the graph reasoning system is a novel realization of self-organized criticality in an AI context. The structural-semantic transition around iteration 400 (see, Fig.~\ref{fig:fig_1}(c)) further underscores the presence of a phase transition-like behavior, consistent with phenomena characteristic of self-organized critical systems.
The observed positive correlation between node betweenness centrality and local semantic neighbor diversity indicates that structurally important nodes tend to connect neighborhoods composed of semantically diverse concepts (Fig.~\ref{fig:fig_3}(c)). Initially, the correlation is strongly positive, reflecting an early phase during which central nodes rapidly integrate semantically distinct clusters. As the network evolves, this correlation steadily decreases and stabilizes at a persistently mild positive value ($\sim 0.15$) around iteration 400, coinciding with the previously identified critical transition. This subtle yet stable positive correlation demonstrates a sustained structural-semantic configuration in which structurally central nodes serve consistently as local semantic bridges, continuously supporting diverse semantic interactions within their immediate neighborhoods. This structural-semantic balance is consistent with the behaviors characteristic of self-organized critical systems. It suggests that the model is still discovering new relationships and further expands novel insights.

Our analysis demonstrates a subtle but consistent semantic entropy dominance, indicating that while structural evolution occurs algorithmically without direct semantic input, it inherently explores a richer semantic landscape implicitly available in the embedding space. This structural-semantic interplay closely parallels critical behaviors observed in natural systems, such as biological networks and phase transitions in physical materials. In particular, the identified critical structural-semantic transition around iteration 400, where entropy cross-correlation shifts from positive to negative, explicitly mirrors physical phase-transition behavior. Initially, structural and semantic entropies evolve synchronously; beyond this critical point, their dynamics diverge, reflecting a deeper intrinsic mechanism by which systems balance structural coherence with semantic novelty.
The stable fraction of surprising edges further elucidates an essential principle: sustained innovation and discovery arise not merely from reinforcing existing structural connections, but crucially through continuously introducing structurally coherent yet semantically novel relationships. This persistent semantic novelty explicitly acts as a reservoir of creative potential, enabling systems to balance structural stability with semantic adaptability and maintain an inherently exploratory, innovative reasoning process.

Ultimately, these findings suggest the existence of universal organizing principles governing both artificial and natural complex adaptive systems. By establishing deep interdisciplinary connections, our results highlight how agentic reasoning architectures, exemplified by the Graph-PRefLexOR model, naturally embody critical phenomena—including subtle semantic-structural interplay, spontaneous structural organization, critical transitions, and sustained exploratory capacity. These insights provide promising foundations for designing next-generation intelligent systems, inspiring interdisciplinary approaches where physics-inspired principles enhance computational creativity, adaptability, and discovery across diverse fields. 

The observation that agentic graph reasoning spontaneously evolves towards a critical state exhibiting semantic entropy dominance mirrors the critical localization transitions originally characterized by Aubry in nonlinear and quasiperiodic systems~\cite{aubry1980analyticity,aubry1997breathers}. Thus, the principles of subtle structural-semantic balance identified here generalize Aubry's seminal insights on critical phenomena into the context of adaptive artificial reasoning systems.

Our central finding reveals that continuous innovation in agentic reasoning systems arises fundamentally from entropy dynamics, specifically a subtle yet persistent dominance of semantic entropy over structural entropy. Structural evolution, measured via Von Neumann entropy, implicitly explores a richer semantic landscape characterized by higher semantic entropy, sustaining structurally coherent yet semantically novel (``surprising'') relationships. This entropy-driven interplay identifies semantic richness as the intrinsic driver of continuous discovery and adaptability, highlighting a universal entropy-based principle underlying complex adaptive behavior in both artificial and natural systems.

In other words, the reason artificial reasoning systems remain continuously creative and innovative may be because they constantly explore a very rich, diverse, and somewhat chaotic space of possible meanings (this is what we call high semantic entropy). In contrast, the actual connections the system forms, its explicit reasoning structure, are more ordered and constrained (low structural entropy). Because the system always has more meaningful ideas available to explore than it explicitly incorporates into its structure, it can continuously discover and create unexpected, novel relationships. This ongoing imbalance between rich semantic possibilities and more structured connections is what fuels sustained creativity and innovation.

Finally, these insights allowed us to propose a new RL framework that uses key insights developed from the experiments conducted in this paper to further tune AI models to become more creative. For such a system to work, the graph-native reasoning model is a naturally fitting framework as it naturally constructs a structured representation that simultaneously captures semantic diversity and relational structure. For this to work with regular text output, we would need to extract or impose a similar graph or latent network representation to calculate these entropy measures.

\section{Methods}

The graphs were generated using the graph-native reasoning algorithm described in~\cite{buehler2025agenticdeepgraphreasoning}, where we let the AI model reason over up to 1,000 iterations. The final graph size has 3,835 nodes and 11,910 edges.
The evolution of agent-generated graphs was analyzed using graph-theoretical and semantic embedding metrics. Structural entropy was computed using Von Neumann entropy derived from the normalized Laplacian of the graph. Semantic entropy was calculated based on cosine similarities of node embeddings.  

\subsection{Node embeddings}
Node embeddings were obtained from a pretrained language model (\texttt{sentence-transformers/all-MiniLM-L6-v2}).  PCA was applied to node embeddings for visualization, revealing structural-semantic relationships and their decoupling. The critical transition in structural-semantic dynamics was characterized by cross-correlation analysis of structural and semantic entropy time series.

\subsection{Entropy Calculation}

We analyzed the structural and semantic evolution of knowledge graphs generated by the Graph-PRefLexOR model, which recursively expands conceptual structures over iterative reasoning cycles. 

Structural entropy was quantified using the Von Neumann graph entropy~\cite{braunstein2006laplacian,kauffman1993origins}, defined explicitly as:
\begin{equation}
S_{\text{struct}} = -\sum_{i}\lambda_i \log{\lambda_i},
\end{equation}
where eigenvalues \(\lambda_i\) are computed from the normalized graph Laplacian:
\begin{equation}
L = I - D^{-1/2} A D^{-1/2},
\end{equation}
where \(A\) is the adjacency matrix, \(D\) is the degree matrix, and \(I\) is the identity matrix. To explicitly ensure rigor and comparability, eigenvalues \(\lambda_i\) were normalized to sum to unity prior to entropy calculation, precisely matching our computational implementation.

Semantic entropy was computed analogously. Node labels were first embedded into a semantic vector space using a pretrained neural language model~\cite{reimers2019sentence} (\texttt{sentence-transformers/all-MiniLM-L6-v2}). A semantic adjacency matrix \(A^{\text{sem}}\) was explicitly computed via cosine similarity between node embeddings \(\mathbf{x}_i, \mathbf{x}_j\):
\begin{equation}
A^{(\text{sem})}_{ij} = \frac{\mathbf{x}_i \cdot \mathbf{x}_j}{\|\mathbf{x}_i\|\|\mathbf{x}_j\|},
\end{equation}
explicitly scaled to the interval [0,1]. Semantic entropy \(S_{\text{sem}}\) was then analogously calculated from the eigenvalues \(\mu_i\) of the normalized Laplacian derived from the semantic adjacency matrix:
\begin{equation}
S_{\text{sem}} = -\sum_{i}\mu_i \log{\mu_i},
\end{equation}
where eigenvalues \(\mu_i\) were explicitly normalized to sum to unity. No thresholding was applied to the semantic adjacency matrix, explicitly preserving the complete semantic relationships inherent in the data, thereby ensuring consistency and rigorous comparability with structural entropy measures.

The correlation between structural and semantic entropies across iterations was characterized by cross-correlation analysis, identifying regime shifts in structural-semantic dynamics. 

\subsection{Community detection}
Louvain community detection~\cite{blondel2008louvain} was used to identify structural communities. PCA projections of embeddings revealed semantic relationships relative to structural communities.  The Euclidean distance of each node from the centroid of its respective community in PCA space was computed. Community centroids were determined as the mean position of all nodes belonging to a given community. The resulting distribution of node distances was visualized using a histogram, where bin colors represent relative node density, with red indicating higher counts and blue indicating lower values.

\subsection{Edge detection and classification}

Surprising edges are defined as structurally present yet semantically distant (cosine similarity $\lambda<0.1$). These were quantified to measure sustained semantic novelty. 

For completeness, we explicitly analyzed semantic entropy sensitivity across semantic similarity thresholds, observing negligible differences at low thresholds ($\lambda\leq$0.2), thus confirming robustness.

\subsection{Correlation between node betweenness centrality and local semantic diversity}
We use the standard definition of betweenness centrality:
\[
\text{BC}(u) \;=\; \sum_{s,t} \frac{\sigma_{st}(u)}{\sigma_{st}},
\]
where $\sigma_{st}(u)$ is the number of shortest paths from $s$ to $t$ that pass through $u$, and $\sigma_{st}$ is the total number of shortest paths from $s$ to $t$.

For each node $u$, let $N(u)$ be its set of neighbors, and let $\mathbf{x}_i$ denote the embedding vector of neighbor $i$. The local neighbor diversity is the average pairwise Euclidean distance among neighbors:
\[
\text{Diversity}(u) 
    = \frac{1}{\binom{k}{2}} 
    \sum_{\,i < j\,} 
    \|\mathbf{x}_i - \mathbf{x}_j\|_2,
\quad
\]
and $k = |N(u)| \text{ is the node degree of } u$. If $k<2$, we set $\text{Diversity}(u)=0$.
We compute Pearson's correlation coefficient between $\text{BC}(u)$ and $\text{Diversity}(u)$ across all nodes to assess how bridging roles relate to semantic diversity in the local neighborhood.

\section*{Conflict of Interest}
The author declares no conflict of interest. 

\section*{Author Contributions}
MJB designed the research, carried out the research, and wrote the paper. 

\begin{acknowledgments}
We wish to acknowledge the support from MIT's Generative AI Initiative. 

\end{acknowledgments}

\section*{Data Availability Statement}

Codes, model weights and additional materials are available at \url{https://huggingface.co/lamm-mit} and \url{https://github.com/lamm-mit/PRefLexOR}. The model used for the experiments is available at~\url{lamm-mit/Graph-Preflexor_01062025}.

\section*{References}
\bibliographystyle{aipauth4-1}
\bibliography{references,references-Mendeley}

\end{document}